# Achieving Operational Universality through a Turing Complete Chemputer

Daniel Gahler, Dean Thomas, Slawomir Lach, Leroy Cronin [*]

*School of Chemistry, University of Glasgow, University Avenue, Glasgow G12 8QQ, UK.*

Email: *lee.cronin@glasgow.ac.uk*

**The most fundamental abstraction underlying all modern computers is the Turing Machine, that is if any modern computer can simulate a Turing Machine, an equivalence which is called 'Turing completeness', it is theoretically possible to achieve any task that can be algorithmically described by executing a series of discrete unit operations[1–6]. In chemistry, the ability to program chemical processes is demanding because it is hard to ensure that the process can be understood at a high level of abstraction, and then reduced to practice. Herein we exploit the concept of Turing completeness applied to robotic platforms for chemistry that can be used to synthesise complex molecules through unit operations that execute chemical processes using a chemically-aware programming language, XDL[7–10]. We leverage the concept of computability by computers to synthesizability of chemical compounds by automated synthesis machines. The results of an interactive demonstration of Turing completeness using the colour gamut and conditional logic are presented and examples of chemical use-cases are discussed. Over 16.7 million combinations of Red, Green, Blue (RGB) colour space were binned into 5 discrete values and measured over 10 regions of interest (ROIs), affording 78 million possible states per step and served as a proxy for conceptual, chemical space exploration. This formal description establishes a formal framework in future chemical programming languages to ensure complex logic operations are expressed and executed correctly, with the possibility of error correction, in the automated and autonomous pursuit of increasingly complex molecules.**

The synthesis of complex molecules can often be translated into a series of procedures that require expert knowledge, continuous observation and attention to detail. On an abstract level,[1-6] such procedures include planning the retrosynthesis of a desired target, conducting a forward synthesis, analysing and purifying reaction mixtures whilst optimising yields and minimising byproducts. For the experienced chemist, these procedures break down into a sequence of well-defined operations which have been standardized and validated by the wider scientific community. Dynamic and responsive procedures are composed of different operations which are then chained together, repeated in loops, and adjusted '*ad hoc*' based on new observations. The chaining together of unit operations[7-10] resembles coded instructions of a computer algorithm. The chemist[11–14] completes each unit, step-by-step, similar to how a compiler interprets a script line-by-line, or a processor executes instructions of machine code one after another[15]. By first defining what a computer algorithm is, one can approach an analogous formalisation and abstraction of a chemical synthesis.

Alan Turing and Alonzo Church laid the foundations for defining the concept of an algorithm in two separate formal frameworks - Turing Machines and Lambda Calculus. Through simulation arguments, these frameworks were later shown to be equivalent, a result known as the Church-Turing-Thesis[16,17]. It is often extended to a third equivalent concept, known as the class of general recursive functions, coined by Kurt Gödel[18]. A common representation of a Turing Machine is a 'Head' positioned along an infinite 'Tape' of discrete cells containing zeros (Figure 1). The Head reads from the Tape, writes to it, and moves to adjacent cells. The Head uses a pre-defined look-up table to determine the sequence of actions, including an update of an internal 'State'. When the Head reaches a pre-defined 'Halt-State', the algorithm terminates, and the execution is complete. Any algorithm can be conceptually reduced into a representation of a Turing Machine with a Tape and a look-up table (though these can become arbitrarily complex). As such, a formal Turing Machine can run any algorithm.[1]

It is useful to explore whether the same algorithmic representation is possible with a traditional chemical synthesis. Unit steps already exist which are combined and built upon to form the foundations of an algorithm, or literature procedure. The chemist, in their role as an experimenter, acts as a probabilistic representation of a Turing Machine, executing these individual operations based on the starting material inputs and the look-up table of similar literature procedures, chemical knowledge and reaction planning within their anthropomorphic head. It is through that human element however, that uncertainty is introduced into the system. Ambiguities in reported procedures, varying skill-sets and inaccuracies following procedures combine to afford erroneous or unpredictable outputs. A deterministic system could be obtained if the human chemist were replaced with an automated synthesis machine that can reliably execute the same unit operations with precision, accuracy and consistency; in practice a robotic system with real-time error correction and fault tolerance would be needed. If this machine was Turing Complete, it could achieve any operation a Turing Machine can, including the execution of constituent unit steps of any algorithm. Consequently, a Turing complete, automated synthesis machine that can perform any unit synthesis, can synthesize any substance that has a known synthesis pathway.

A method of precisely encoding the synthesis pathway of a chemical is critically important for an automated machine to execute reliably. The Chemical Description Language (XDL) is a concept designed to unify abstract chemical operations into machine-readable and executable base steps, which allows for the automatic execution of chemical syntheses in a non-ambiguous way[7]. By compiling the unit steps into new forms, XDL adapts to permit more elaborate reaction types, complex procedures, and increased programmable features. Fundamentally, XDL is hardware-independent, as each platform can map the respective bindings to XDL in order to execute the chemical syntheses[8–10,19]. Examples of such platforms are the ChemPU, OpenTrons, RoboticArmXDL and Baristabot[20,21].

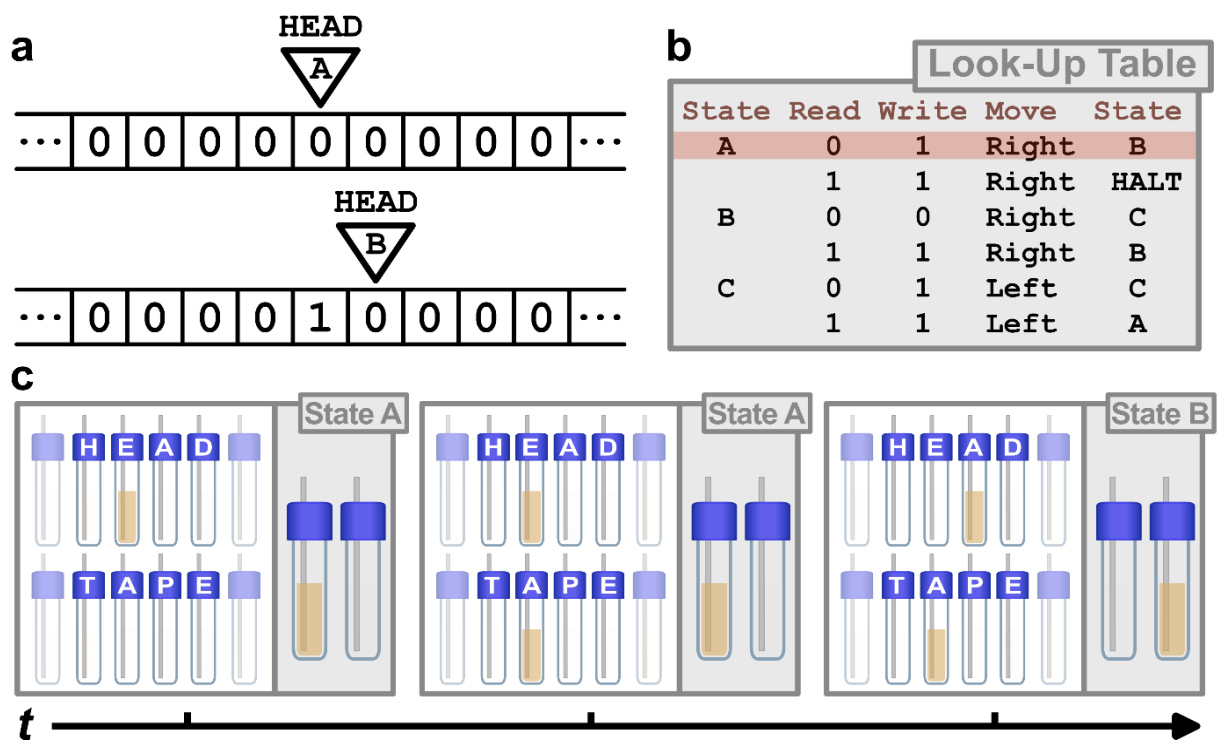

**Figure 1** – Turing Machine Representation. [a] A traditional and simplified representation of a Turing Machine comprising of a Head in State A reading along a theoretically infinite Tape then writing information before changing State to B. [b] An exemplar look-up table corresponding to the algorithm. [c] An illustrative representation of the same Turing Machine and its operation utilising the colour of solutions in filled vials as a proxy for reading and writing data.

The hardware-independence of XDL permits chemical reactions to be captured in an abstract form that can be executed on any chemical platform with sufficient hardware modules. A simple reaction that consists just of additions, heating and stirring can be executed on almost any chemical platform as these are elementary chemical steps. A more elaborate reaction featuring a separation, evaporation or column-chromatography can only be executed on platforms where corresponding hardware exists. If the required hardware does not exist however, it can still be possible to write the XDL file, using the current standard[22].

In order to achieve Turing completeness on an automated synthesis machine such as the Chemputer, we closely follow the paradigm of simulation arguments from the proof of the Church-Turing thesis (Figure 2). Given two systems A and B, we say system A can *simulate* system B (or B can be simulated by A), if every function of system B can be completely and deterministically represented in system A. We abbreviate this with the notation A→B. In the Church-Turing thesis, there are two systems, as shown in Figure 2a. The Turing Machine can simulate Lambda-calculus, Turing → Lambda ($a_1$) whilst Lambda-calculus can simulate the Turing Machine, Lambda → Turing ($a_2$)[17]. This simulative property is necessarily transitive: meaning two simulations of systems A→B and B→C can be chained to get a simulation A→C (See SI 2.1.6 for examples). To demonstrate that XDL is Turing complete (Figure 2b), it is needed to be proven that:

- The Turing Machine can simulate XDL, Turing → XDL ($b_1$ and $b_2$)
- XDL can simulate the Turing Machine, XDL → Turing ($b_3$)

XDL represents an automated synthesis machine, such as the Chemputer, executing abstract procedures captured in XDL (See SI 2.1.5 for details). The Turing Machine represents the abstract concept. From Hopcroft[1,23], we take that the Turing Machine can simulate the computer, Turing → PC ($b_1$). In fact, Hopcroft even shows that there is an arrow in the opposite direction PC → Turing ($b_{-1}$), in other words, modern computers are themselves Turing complete. Next, representing a simulation of XDL in a computer, PC → XDL ($b_2$). This is possible and is adequately called "simulation mode". This simulation mode is used in practice to perform final checks before executing a XDL procedure. By transitivity, the two statements, Turing → PC ($b_1$) and PC → XDL ($b_2$) combine to Turing → XDL. What remains to be shown is the final connection, that XDL can simulate the Turing Machine, XDL → Turing ($b_3$). If that last step is closed, it would demonstrate that the systems XDL and Turing Machine can simulate each other. That is the definition of Turing completeness, and thus concludes the proof.

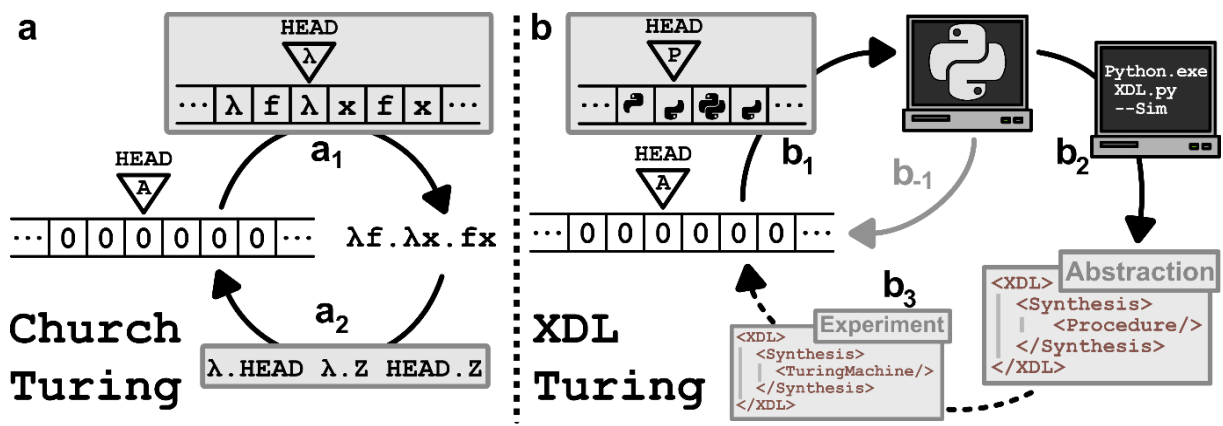

**Figure 2** – Illustrative Proofs of Turing Completeness. [a] The abstract concepts of the Turing Machine (left) and Lambda-Calculus (right). $a_1$ implies the ability of the Turing Machine to simulate Lambda-Calculus, $a_2$ implies the converse. [b] The abstract concept of the Turing Machine (left), a modern computer running python (top) and XDL running on a ChemPU (bottom-right). Each arrow implies the ability of its source system to simulate the destination system. $b_1$ is the simulation of the python computer in a Turing Machine which follows from Hopcroft. $b_{-1}$ is the simulation of a Turing Machine in Python, which is a common coding exercise. $b_2$ is the simulation of a XDL procedure purely in python, called simulation mode in XDL. $b_3$ is the simulation of the Turing Machine in XDL which needs to be demonstrated.

To realise an automated synthesis Turing Machine in the physical world, firstly the theoretical concept has to be simplified and the limitations of finiteness and size restrictions must be accepted (see SI, Section 2.1.4). Secondly, XDL needed to incorporate conditional execution as a feature that enables Turing completeness and thus elevating into a programming language. The introduction of conditional execution was established as follows: To execute a step based on whether some condition C is true, in analogy to "If (C), Then execute Step S1, Else execute Step S2", the argument "condition" is introduced to the (set of) steps S1 that should only be executed if that condition C is true, i.e. condition="C".

```
<S1 parameter_1="value_1" … parameter_n="value_n" condition="C" />
<S2 parameter_1="value_1" … parameter_n="value_n" condition="not C" />
```

This condition C is either the result of a comparison (for example, is the mixture in reactor1 red?) or a Boolean expression whose smallest units consist of such checks (for example, is the mixture in reactor1 red AND the temperature in reactor1 below 50° AND is the pH in reactor1 between 4 and 9?). Therefore, what remains is to define a method that executes such a comparison and stores the result of said comparison in a way that it can be used as part of the condition C. A Measure step is therefore introduced, which has a mandatory unique step_id and generates a truth-value from such a comparison, based on sensor data. We can, for example, at the appropriate moment during the reaction, execute a Measure step with the step_id "C", the measured quantity "colour", the comparison to "equal" and the value "red".

```
<Measure step_id="C" target="reactor_1" quantity="colour"
comparison_value="red" true_if="equal"/>
```

If, at the time of the execution of the step, the colour is measured to be red, the pair of data {C:true} is stored (similar to how Boolean values are stored in variables in any other programming language). If it is not, the pair {C:false} is stored. This can then be used as a condition in a step, or in combination with others as part of the condition of a step. One such usage within a chemical application would be to decide whether to quench a reaction based upon a relevant variable, for example the colour of the solution (Figure 3). A discussion on why this was previously not possible is found in SI 2.3.3.

For this new version of XDL to be considered Turing Complete, two arguments must be fulfilled. The first is to show that this method allows for *any* kind of conditional execution, the second is a proof of work and that conditional execution holds in practice. Using theoretical arguments is not a clean way of proving this, as XDL is rather complex. Instead, the commonly accepted way of proving Turing completeness is through a simulation argument. We show here that the new XDL can simulate any Turing Machine, concluding the proof of Turing completeness.

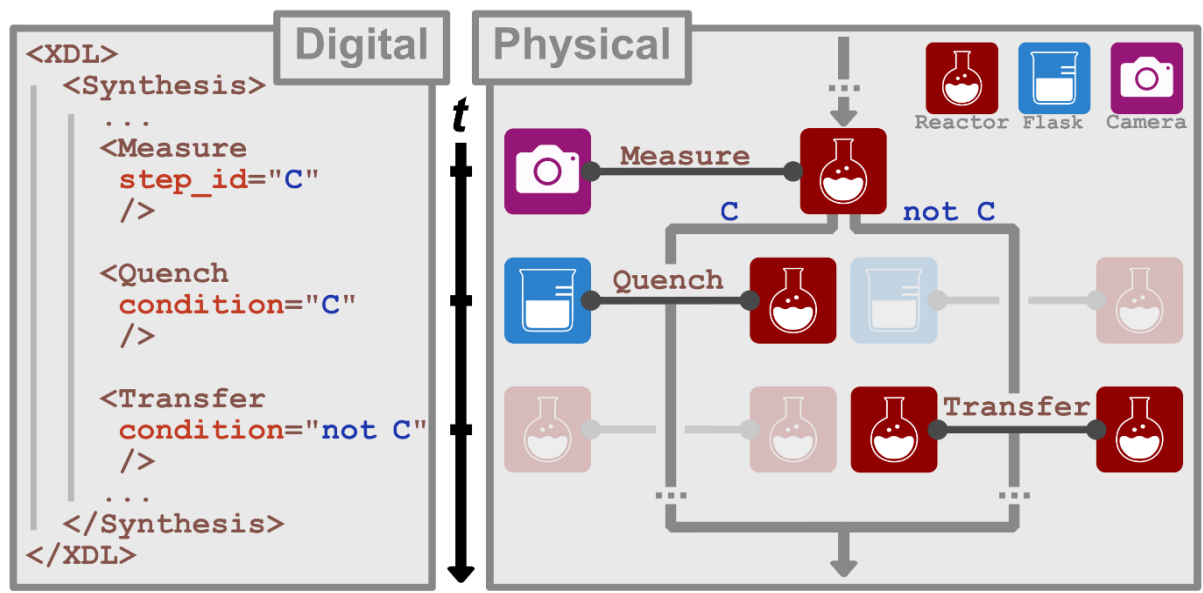

**Figure 3** – Application of conditional execution. Left: digital representation in the file. During the procedure, a Measure step is executed storing the result of the measurement in the variable called C. Quench is executed if C is true, but not if C is false. Transfer is executed if C is false, but not if C is true. Right: physical representation in the lab. After the measurement, two paths can be taken, based on the measurement. If C is true, the quench step will be executed, the transfer step will be skipped. This is represented by the left path. If C is false, the quench step is omitted, and the transfer step is executed. This is represented by the right path.

In order to simulate a theoretical Turing Machine in the real world, we need to represent all parts in hard- and software, in particular, the Head, State, Tape and alphabet. This was emulated using a Chemputer (Figure 4a). The platform consists of three pumps, five valves, three sequences of vials and a camera (Figure 4b). The bottom left sequence of vials represents the Tape. The sequence of vials above represents the position of the Head. The two vials in the bottom right signify the state. The alphabet is defined by the presence and colour of liquids in the vials. In this example, the alphabet consists of four letters, represented by orange, blue and green solutions, as well as an empty (white) vessel. These four colours are also permissible for the state-vials, however the Head vials will only ever have one orange vial signifying the position of the Head, with the other seven being empty.

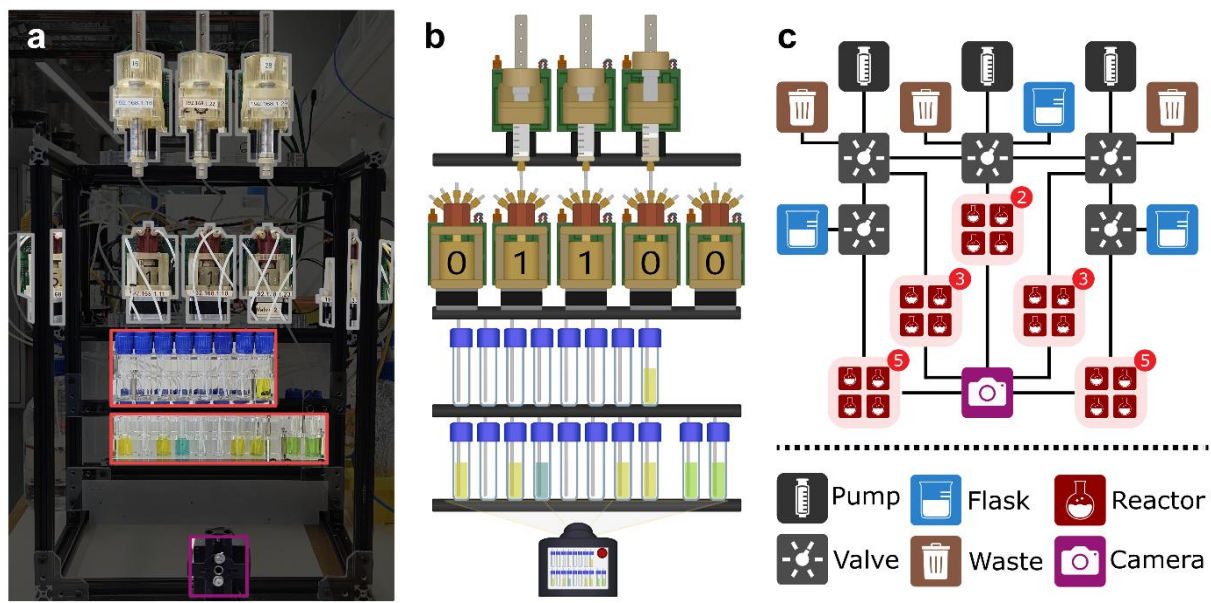

**Figure 4** – A Turing Machine Simulated in a Chemputer. [a] Physical platform used to run the experiments. Highlighted in grey are the pumps and valves which are the components of the Chemputer's liquid handling backbone. In red is the sequence of vials representing the Head, Tape, and two vials indicating the state. Highlighted in purple is the camera. [b] Simplified representation of the core objects of physical platform. [c] Digital representation of the objects in the ChemPU, similar to its representation in the ChemIDE[24].

With the physical and graphical representation established, we now construct the Turing Machine in software as a sequence of nested XDL blueprints. A blueprint in XDL is the equivalent of a function in programming and a simplified representation of blueprints is illustrated (Figure 5). For a diagram of the full nested structure (see SI 3.2). Since a Turing Machine continuously loops until the Halt state is reached, the initial outer blueprint "TuringMachine" consists of a Repeat step with an exit condition of "not HALT", meaning it will run until the HALT state is reached. Within the loop, the state and tape are read, and the required action is taken from the look-up table. These are encoded in the blueprints "ReadState", "ReadTape" and "LookUpTable". Within blueprints such as ReadTape, a measurement is made to determine the colour of a vial and then interpret that as a value for the Turing Machine. This is established through the Measure step, which allows us to store Boolean values with their step_ids. The LookUpTable blueprint then uses the values from the

step_ids to determine which of its substeps to execute. Shown here, the actions from the look-up table of "write 1, move right, switch to state B" are themselves encoded as blueprints "WriteOneMoveRight" and "SwitchToStateB". Necessarily, at least one of the options in the look-up table will switch to state HALT to ensure there is a chance of the algorithm terminating, though that is not guaranteed, a problem known as the Halting problem[25,26].

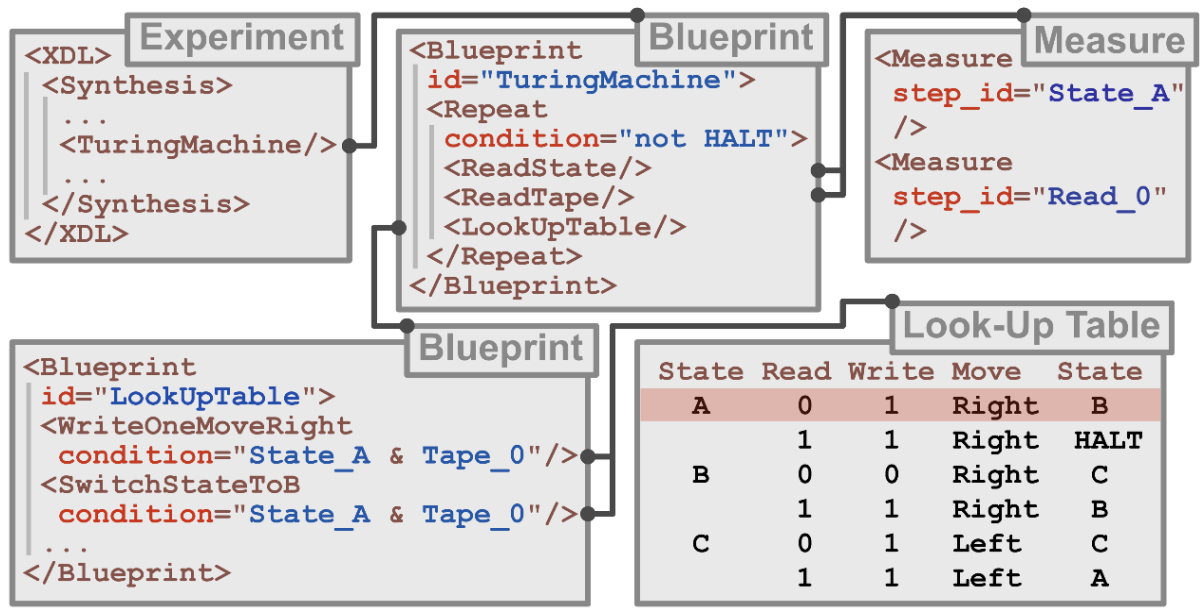

**Figure 5** – XDL representation of the Turing Machine. The XDL equivalent of a function in programming is called a blueprint. Top-Left – The original XDL file calls the TuringMachine blueprint, similar to how a compiler would call a function that executes a Turing Machine. Top-Middle – That function/blueprint is a Repeat-loop, which iteratively obtains the state and the label on the Tape position. If the state is anything other than "HALT", it executes the instructions from the lookup table. Top-Right – The Measure Step is used to determine the current state and the label/contents of the Tape in the current position. The information is stored in Boolean variables. Bottom-Right – The lookup table is consulted with the stored data. The two leftmost columns determine which row of the table to use, organized by State and Read label. In this example, State is A, Read is 0. The other three columns show the actions to be taken, in this example: write a 1 in the current position, move the Head to the right, switch to State B. Bottom-Left – The lookup table in a XDL blueprint. It consists of a triplets of steps: Write, Move, and Switch, which will be executed based on the conditions (State being A, label being 0 being true).

With all the parts in place, we now demonstrate the example experiments. For each experiment, the blueprint containing the look-up table is adjusted to reflect the information of the given algorithm, the Tape, Head and State are initialized to their starting values, and the main XDL file is executed (See Figure 6 and Supplementary Videos 1-3 for a visual representation of the execution of the algorithms).

Experiment 1: Busy Beaver

A well known example and problem of Turing Machines is determining how many consecutive ones you can write on an empty Tape, given the number of states n. This function is known as the Busy Beaver function $\Sigma(n)$, and in our first example, we will show $\Sigma(3) \geq 6$ by using 3 states to write 6 consecutive ones. These six rules, visualized from the look-up table (Figure 6a), consist of two rows: on the top, the position of the head is signified by the label of its state, and the background colour (orange/white) signifies the Tape value it is reading (1/0). On the bottom, the action is described. The new state is signified by the new label, the position of the state signifies the moving direction, and the colour in place below the original state signifies whether to write a 1 or 0. For example, the top right rule has in its first row the label A on a white background, describing the situation where the head is in State A and reading a 0. The row below has a label B to the right, and is orange below the original state. This indicates that the Turing machine will write a 1, move to the right and switch to state B.

Experiment 2: Binary Addition

While discussing the basics of computation, a good example of a simple program is the addition of two binary numbers. Given two binary integers of size 3 bits each, we wish to calculate their sum. We extend the alphabet to the symbols "0", "1" "x" and "y", and for the initial integers "abc" and "def" where a,b,c,d,e,f represent the bits of the two numbers, we initialize the Tape with the sequence "abcx0def". For example, let abc=101 the number 5, and def=011 the number 3. We initialize the Tape as "101x0011" and expect the output 3+5=8, in binary 0001000. Further details and discussions can be found in the Supplementary Information, Section 2.5.3.

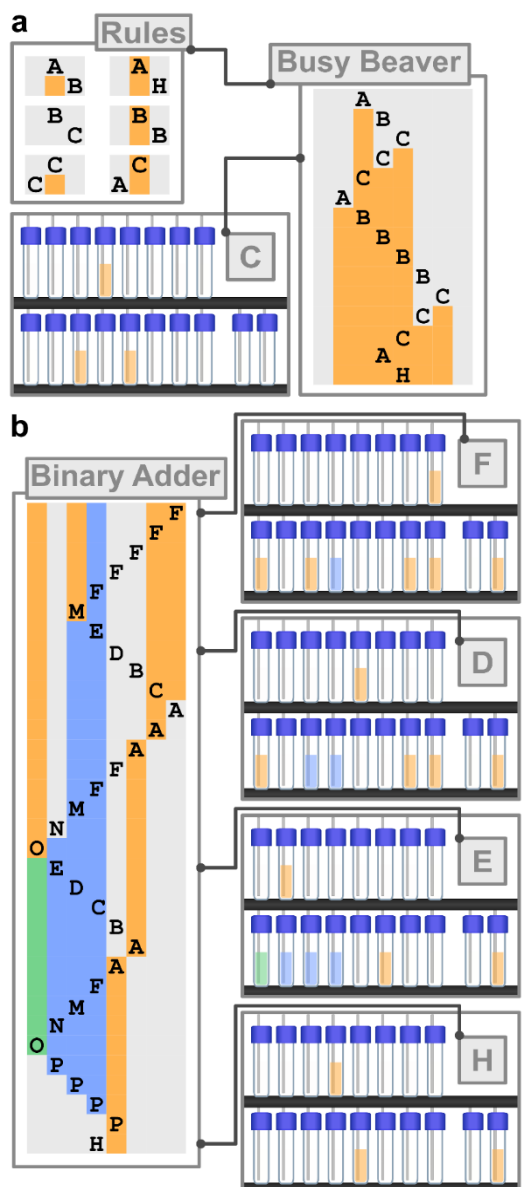

**Figure 6** – Algorithm Execution on the ChemPU[27]. a – Busy Beaver with schematic representation of the look-up table as individual rules. For example, the first rule is to be read as: The Head is in state A and the Tape is white (as the background behind the letter A is white). The instruction is to write "orange" to the Tape, move to the right, and switch to state B (As the colour underneath the letter A is orange, the letter in the second row is a "B" and is to the right of the initial field. It corresponds to the first row of the lookup table in Figure 4 (Bottom-Right). Middle: Execution from top to bottom. Each line represents one execution step of the Turing machine, with the first line being the initial empty Tape with the State being "A" and the Head on the third vial, represented by the position of the state "A", and the last line being the result in the Halt state with six consecutive orange vials representing six ones in a row. Bottom: representation of the entire vial configuration three steps from the bottom. The Head is on the third vial from the right, as this is the full vial in the top row. The state is "C" by the encoding convention of the two vials in the bottom right. The Tape reads 01111010 represented by orange and empty vials. b – Binary Adder. Left: Execution from top to bottom. The first row represents the example setup of 101x0011 for the addition of 101 and 011. The bottom row represents the result of 00001000 yielding the result of the binary addition 101 + 011 = 1000. Right: Representations of the entire vial configuration at several moments during execution. See supplementary videos for full explanation of individual states

Conditional logic, when integrated with automated synthesis machines, holds significant potential to advance the state-of-the-art in chemistry. Traditional, manual experiments typically monitor a reaction for a fixed duration, concluding when the allotted time has elapsed, regardless of real-time changes to the reaction conditions (Figure 7). This rigid approach often limits optimization and adaptability, preventing scientists from making dynamic adjustments that could improve yields or overall outcomes. By contrast, smarter systems equipped with conditional logic can revolutionize this process. These systems can monitor reaction parameters such as colour, quenching temperature and yield in real time, only proceeding to the next step once a target or plateau has been achieved. Such adaptability transforms synthesis from a linear, time-based process into a dynamic, multi-dimensional pathway. This allows for repeated adjustments and reconfigurations that continually optimize the reaction conditions, yielding better results with less human intervention. Moreover, sub-routines can be integrated to address errors and ensure greater reliability. For example, if an error is detected during the reaction process, the system could automatically restart or recalibrate the platform, minimizing downtime and human oversight..

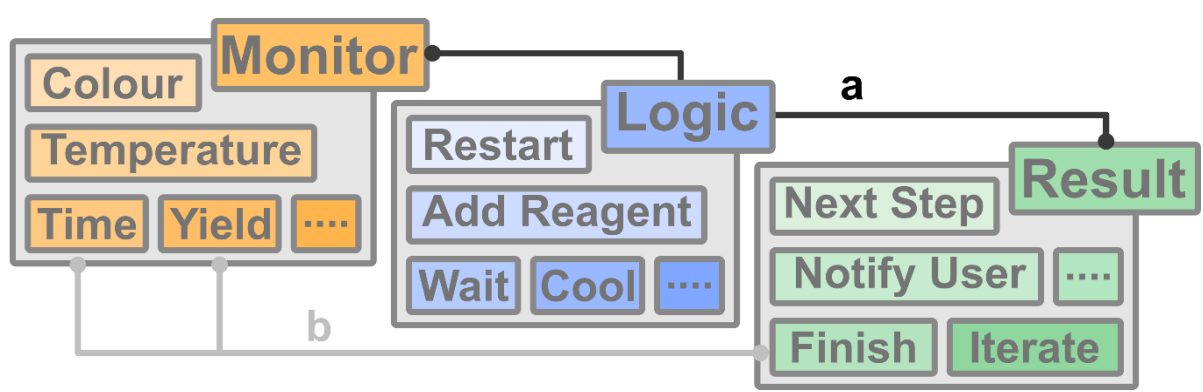

**Figure 7** – Flow diagram representing the rapidly growing potential of Turing-complete, automated synthesis machines which incorporate conditional arguments (a) to augment fixed procedures dynamically. This approach contrasts traditional methods (b) which utilise time as the only end-point variable, or in more advanced systems, on-line yield determinations.

This automation enhances both the safety and reliability of chemical synthesis, reducing the risk of accidents and ensuring more consistent results across experiments. Consequently, these advancements pave the way for more robust, intelligent, and efficient chemical production methods, addressing key challenges in modern chemistry.

As complex as chemical syntheses are to an external observer, there are rules which allows the chemist to subdivide most reactions into a series of unit operations such as Add, Heat, Stir and Wait. These steps can be communicated with other chemists and are widely understood. If all procedures are diligently documented, a digital chemist can capture a linear synthesis pathway into a computer algorithm to be executed by common automated synthesis machines with precision, consistency and reliability. There are however, reactions that require complex, conditional arguments and nested decision loops in order to achieve the best outcomes. As such, these pathways cannot be easily captured into algorithms and executed on common synthesis platforms. We have shown that with our critical development on the XDL codebase, the Chemputer became a Turing-complete synthesis machine, that can run any algorithm in XDL. Consequently, any molecule with a known synthesis pathway can be synthesized with the Chemputer, a property that can be called chemical Turing completeness (Turing Chempleteness)[28]. This property will be inherited by any synthesis machine with sufficient hardware running on XDL.

**Supplementary Information**

Extended definitions, methods, XDL blueprints and procedures and are available in the accompanying Supplementary Information (SI) files, along with Supplementary Videos (SV) that demonstrate the operation of the Busy Beaver algorithm on a small (SV1) and large (SV2) scale alongside a Binary Adder algorithm (SV3).

Video Captions:

SV1 - Busy Beaver on two states A and B, as well as the halting state H, writing three consecutive ones. The upper row of four vials represents the head, the lower row of four vials represent the tape and the two rightmost vials on the bottom row represent the state.

SV2 - Busy Beaver on three states A, B and C as well as the halting state H, writing six consecutive ones, as seen in Figure 6a. The upper row of eight vials represents the head, the lower eight vials represent the tape and the two rightmost vials on the bottom row represent the state. For further details, see SI: 2.3.1

SV3 - Binary Adder of two, three digit numbers, in the example of $5 + 3 = 8$, represented in Figure 6b. The numbers 5, represented by 101 on the left of the blue vial, and 3, represented by 011 on the right, the result being represented by 1000. The upper row of eight vials represents the head, the lower eight vials represent the tape and the two rightmost vials on the bottom row represent the state. For further details, see SI: 2.3.2

**Acknowledgements**

We gratefully acknowledge financial support from the EPSRC (grants nos. EP/L023652/1, EP/R020914/1, EP/S030603/1, EP/R01308X/1, EP/S017046/1 and EP/S019472/1), the ERC (project no. 670467 SMART-POM), the EC (project no. 766975 MADONNA) and DARPA (projects nos. W911NF-18-2-0036, W911NF-17-1-0316 and HR001119S0003). The views, opinions and/or findings expressed are those of the authors and should not be interpreted as representing the official views or policies of the Department of Defence or the US Government.

**Author Contributions**

LC conceived the idea of making XDL Turing complete and proposed a formal abstraction. DG created the implementation, coding abstraction and proof, with feedback from LC. LC and DT coordinated the research project. Software development was completed by DG and SL with help from DT. The manuscript and the SI were written by DG, DT and LC.

**Competing interest statement**

The authors declare that they have no competing financial interests.

**SUPPORTING INFORMATION**

**Achieving Operational Universality through a Turing Complete Chemputer**

Daniel Gahler, Dean Thomas, Slawomir Lach, Leroy Cronin*

School of Chemistry, University of Glasgow, Glasgow, G12 8QQ,

# Contents

# 1. Materials and Instrumentation

## 1.1. Materials

The reagent-grade chemicals were obtained from Fluorochem, Sigma-Aldrich and TCI. Other reagents were used as obtained without further purification. Solvents were purchased from several departmental suppliers, Honeywell, Fisher and Sigma-Aldrich.

## 1.2. Chemputer Platform

The pumps, valves and frames are standardised pieces of equipment designed by the Cronin Group and assembled as required. Fittings, adaptors, tubing and other commercially standardised parts are purchased from suppliers including RS Components. The exact part numbers for a standardised platform alongside extensive build instructions are readily available[1].

## 2. Technical Discussions

### 2.1. Turing Machines

**2.1.1.** Clarification to Abstract "**The smallest common unit that every modern computer breaks down to is called a Turing Machine**": There is no physical head, tape and state register at the lowest level of actual computers. They have however been proven to be equivalent, see further discussion later in the paper.

**2.1.2.** Clarification to Introduction "a formal Turing Machine can run any algorithm": Any algorithm in the modern sense can be expressed as a Turing machine which can be executed. However not any Turing machine will reach the Halt state, therefore not every such execution will finish and end with a readable result on the tape. This is not a limitation, but an exact analogy, as halting/non-halting is a property of the algorithm itself. The problem on whether a given Turing machine will halt is known as the Halting problem and was already known to Turing to be undecidable. When we say "can run any algorithm" we mean the execution itself, not the Halting.

**2.1.3.** Generalization of the Halting problem: To be more precise, the halting problem is semi-undecidable. What that means is that to a given Turing machine we cannot decide whether it halts or not UNLESS it halts, in which case we can observe that it halted and hence determine that it halted. This can be extended to the problem of whether a certain program executes a function with a result. This problem is semi-decidable in the exact same sense and for the exact same reason. If a program is observed to execute the function with the result, then it can be decided that it does in fact do that. However, if we do not make such an observation, we CANNOT determine that the program does NOT do that. Furthermore it is semi-decidable whether a program will perform a specific task or return a specific output. If it does, it is decidable that it does, but showing that it doesn't is undecidable.

**2.1.4.** On Finiteness: Since we assume that the number N of atoms in the universe is finite ($\sim 10^{80}$), no actual physical turing machine with an infinite feedtape can exist. We therefore subscribe to the concept of a generalisation of the finite machine: We can build and simulate machines with finite tapes. Every program that halts (see 2.1.2, 2.1.3) will only require finite tape to execute. Therefore for a given maximal execution tape length n, even if n > N, as long as we have a method to build a machine if we *had* enough atoms, we will agree that we were able to simulate that machine. For example, assume N to be $10^{81}$, and we want a Turing machine that generates the $(10^{81} +1)^{th}$ prime number p in unary, we know p > N, and therefore cannot run the program within our limited universe. However we can think of a procedure to extend a computers memory assuming we had enough material to create it, up to the point to where it would be able to print out p. Since every such program would only require a finite tape, it would require a finite amount of atoms, even if more than N are required, we can assume that we can simulate the machine. In short, similar to complete induction, we simulate a finite case, and explain the extension to larger tapes. In particular, it does not matter where we draw the line of how much physical memory our instantiation of the Turing machine has. It will limit the number of machines we can run, but this limitation will always exist in the real world.

**2.1.5.** The proposition "XDL is Turing complete" is specified as follows. Whenever we say XDL in this context, we mean the language together with an implementation and an execution platform. While we specifically refer to a specific Chemputer consisting of a PC connected to an automation rig running XDL, the resulting statement will hold true on any platform with a sufficient implementation of the simulation more and sensor control.

**2.1.6.** Examples for simulations, formal definitions:

- It is possible to code up Conway's Game of Life with a few lines of code on a modern computer. When that code is running, the computer together with that code simulate Conway's Game of Life.

- Virtual machines[2] can simulate entire operating systems within each other, like a VM running Linux Mint on a Windows machine, and vice versa.

- The set of ducks entering and leaving a pond simulates the natural numbers with incrementation and decrementation.

- By transitivity, Linux simulating Windows simulating Conway's Game of Life implies Linux simulating Conway's Game of Life.

A precise definition of "simulation" is difficult as it slightly different meanings depending on the context. We use it in two ways that are similar, but not identical. First, in the conceptual discussion of abstract mathematical concepts, where it just means that one mathematical object satisfies the definition of another. This is what Turing used in to prove the equivalence of Lambda Calculus and Turing Machines[3]. Later we use it in the context of physical hardware (e.g. a computer running python) simulating systems in a computer program. While you can interpret the python script itself this in the previous way, i.e. the construct represented by the code satisfies the definition of the target object (e.g. a script that produces a Turing machine), you can also run this code in a physical location and execute the simulation in a real sense.

## 2.2. Chemistry

**2.1.7.** Clarification to Introduction "we minimize or eliminate the randomness introduced by the human and obtain a deterministic system": Every real life measurement is prone to error. While we eliminate the potential sources of uncertainty introduced by chemists doing manual synthesis, chemistry itself is rarely considered to be fully deterministic, a discussion that will break down in the very latest on the level of observations on a quantum-mechanical level.

**2.2.** XDL and Chemputers

**2.2.1.** It is important to distinguish between XDL and the ChemPU: XDL is the abstract, hardware-independent language, the ChemPU is an entire system as a sequence of bindings from XDL down to the hardware.

**2.2.2.** Clarification of Introduction "…XDL needed to incorporate conditional execution as a feature that enables Turing completeness and thus elevating into a programming language": It is important to note that XDL as of 2024 does not constitute a programming language per se, similar (yet not identical) to how HTML is not a programming language, since neither of them are Turing complete. There are (thankfully) several ways of describing Turing completeness as a concept, and it can be easily described what certain languages are missing to become Turing complete. Very often, as in this case, it boils down to the missing of conditional jump instructions, also known as branching or "If/Else" statements, that then allow for unbounded iteration, such as the construction of "While/Until" loops. On a Hardware level, for example in von Neumann architectures[4], this is established by NAND-Gates that enable conditional jump instructions in the machine code. HTML does not have these, but JavaScript/ECMAScript does, and as such IS Turing-complete. XDL itself also does not have If/Else statements, which we will henceforth call "conditional execution". While it *does* feature unbounded iteration of some cases, like the DoUntil step or the Monitor step, which allow the execution of code until a measurement reaches a certain threshold, it is not constructed from conditional execution, as that is not a feature of XDL. To understand the difference, consider the following example. After a certain reaction, the mixture in a reactor is either red or it isn't. If it's red, we need to add reagent A, if not, we don't. This cannot be expressed in XDL as of 2024 (This constitutes a halting problem, see Section 2.1.3.)

**2.2.3.** Comparison with older version of XDL before conditional Execution: Going back to the previous example that cannot be expressed in old XDL, this can now be expressed in the new XDL: After the reaction, a Monitor step is executed, with the ID "colour_red" that compares the colour of the reaction mixture to a predefined value of red with respect to some error (this of course is fuzzy and dependent on the errors, camera, and image processing among other things, however it is shown in chapter 3 that these restrictions are in fact of little issue). If the colour is identified as red, XDL associates the value "true" to the ID of "colour_red", otherwise it associates the value "false". Afterwards, an Add step of reagent A, with the condition "colour_red" is set. It is executed, if and only if "colour_red" evaluates to "true". That happens if and only if the colour of the mixture was identified as red. In conclusion, Reagent A is added if and only if the reaction mixture was red, as required. For this, we required the implementation of a Camera module to the ChemPU, which is released together with the new XDL.

**2.3.** Experiments

**2.3.1.** Busy Beaver: As a motivation, we can ask ourselves, what is the largest finite number we can generate with 10 characters of Python code [possible answer: "9<<(7<<33)"]. Theoretical coding exercises such as these are known as code golf. One of the earliest such exercises was popularised by Rado in 1962[5], where the question was, given a limited alphabet and number of states, what is the longest sequence of "1" on an initial empty tape that a Turing Machine can write. The question leads to a surprisingly fast-growing function. Denote by $\Sigma(n)$ the number of sequential "1"s for n states. Given a binary alphabet of "0" and "1", there are in fact currently only three known values: $\Sigma(2)=4$, $\Sigma(3)=6$, $\Sigma(4)=13$, the rest are estimates, $\Sigma(5)\geq 4098$, $\Sigma(6)>10\uparrow\uparrow 15$ (using Knuth's arrow notation). In the experiment, we demonstrate $\Sigma(3) \geq 6$. Note that in order to prove equality, one must show that no other Turing machine with 3 states generates a larger sequence. It is due to this requirement of complete enumeration that we only have estimates for larger values of n.

**2.3.2.** Binary Addition: The addition of binary numbers is a simple yet essential part of the inner workings of computers, we assume the reader to be familiar with the concepts of carry-in and carry-out. We explain the setup of the Turing machine and look-up-table and explain the execution of the example in Figure 6b and the video file. We add a 3 bit binary number to a 4 bit binary number. The alphabet consists of 4 colours: white, orange, blue and green. As inputs, the numbers are represented on the tape with an orange vial for a 1 and a white vial for a 0. The second number is separated from the first with a blue vial. The head starts in the rightmost position, in state F. The purpose of state F is to "Fetch" the next non-zero bit of the left number, and its instructions are to always move left, write what it reads, and if it reads the blue vial, to switch to M. State M is the first of the three states M, N, O that will initialize the transport of the digit back to the left. If the head is in one of these states and reads white, it continues to the left and switches M -> N --> O. If it reads orange, it continues to the right and switches to E. M and N will write blue, whereas O will write green, signifying the final digit of the left number. If O reads white or green, this indicates that all digits have been added and it will initiate the purge, by switching to state P. Assume that one of these three states, M, N O finds an orange vial, moves right and switches to E. E is part of the states that transport exactly 5 steps to the right, by switching E->D->C->B->A, moving right and writing what the head reads. When A is reached, the actual addition begins. The state A signifies the addition of a 1 in the current position, which can be due to either adding a bit from the left number, or by carry-in of a previous addition. If it reads white, it executes 1+0=1 with no carry-out, and therefore writes orange, moves left and switches back to F to fetch the next number. If it reads orange, it executes 1+1 = 0 with a carry-out, and therefore moving to the left, remaining in state A, thereby carrying the previous carry-out to the next carry-in. When all digits from the left have been added to the right, the head in the fetch state will advance to the leftmost vial in the O state, and will either read white or green, in which case it will switch to the purge state P, write white and move to the right. P itself will also write white and move to the right, unless it reads orange, at which point it will have reached the leftmost digit of the sum. It will then move to the left, switch to the halt state H and terminate. We can see this executed in the example in Figure 6b by following the flow line by line. We add the numbers 5 and 3, represented as 101 and 0011, therefore start the tape with 101x0011, with 1 representing orange, 0 representing white and x representing blue. The head begins on the leftmost vial in state F, and proceeds to move left to fetch a digit, without changing the tape. On the blue vial it switches to M, which reads orange, overwrites it with blue and begins the transport to the right by switching to E and moving right. E switches down all the way to A without changing what's on the tape. A then adds the transported digit to the leftmost vial, resulting in writing a 0 and carrying a 1. It moves to the left, remains in state A. It again reads a 1, writes a 0, moves to the left and stays in state A. Finally it reads a 0, writes a 1, and moves left to fetch. Again, on blue, F switches to M. M is responsible for the rightmost digit of the left number, which has already been taken care of. It reads blue and switches to N, and moves on to the left. N is responsible for the middle digit of the left number, which is zero and therefore ignored. It switches to O and moves to the left. O reads a final orange, writes green and begins the transport to the right. Again, E turns all the way to A which adds the digit exactly in the right position. It performs a carry by writing a 0 and moving to the left, where it writes the final 1. A does not know that it is over, and switches back to fetch. F will reach M, N and O, which then reads green and initiates the purge. It writes white, switches to P and moves left. P does the same, and overwrites all blue vials with white. It stops when it reads orange, switches to H and moves to the left. The program terminates and the output is the tape 00001000, which is the binary representation of the sum 8=3+5.

## 2.4. Proof of Turing Completeness

**2.4.1.** Transitivity of Simulations: Given three systems, A, B, C, A can simulate B (Statement 1), and B can simulate C (Statement 2), we want to show that A can simulate C, which means that every function of system C can be completely and deterministically represented in A. So take any function f of system C. By statement 2, this function can be completely and deterministically represented by a function in system B, and call that function g(f). By Statement 1, we can represent any function g of B deterministically and completely in system A, call that function h(g). Combining both of these, we get a function h(g(f)) in A that completely and deterministically represents the function f in C.

# 3. Chemical Turing Machine

**3.1.** The following describes the full flow of the Turing machine blueprints as used in the code that ran in the lab. The main loop was the TuringMachine blueprint that measured/monitored the HALT state and had an ERROR catching mechanism. It would loop through the Transition blueprint continuously until HALT or ERROR received True. The Transition blueprint begins by resetting all variables, then reading the state from the state vials and the tape by iterating through the head vials and reading the tape vial where the head vial was non-empty. The Transition blueprint would then enter a sequence of conditional blocks that set the colour to write, the direction to move, and the state to switch to. This is designed in a way that exactly one of these blocks are hit during execution, encoding the look-up table that corresponds to the given Turing machine. This is the only part of the blueprint structure that needs to be adapted when switching to a different Turing machine, as all information on the algorithm is contained in the look-up table. Finally, the tape is written to, the head is moved, and the state vials are updated, before the main loop in TuringMachine starts anew.

## 3.2. Diagram of the full Turing Machine blueprint as ref

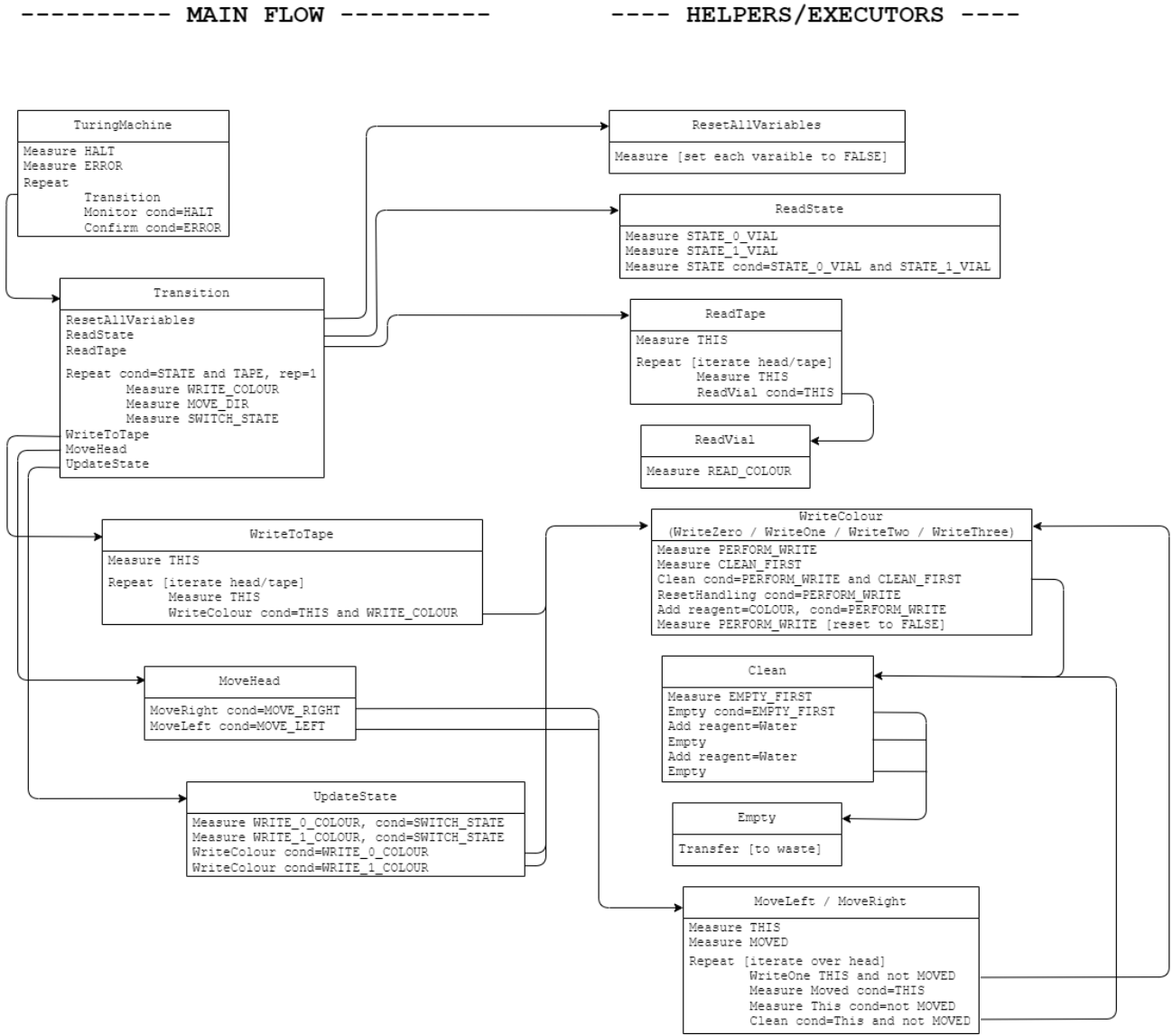